\documentclass[10pt,twocolumn,letterpaper]{article}

\usepackage[pagenumbers]{iccv} 

\usepackage{multirow}
\usepackage{multicol}
\usepackage{array} 
\usepackage{amssymb}
\usepackage{pifont}
\def\etal{\emph{et al}.}
\def\ie{\emph{i.e.,}}
\def\eg{\emph{e.g.,}}
\def\sota{\emph{state-of-the-art}}
\newcommand{\winner}[1]{{\textbf{#1}}}
\newcommand{\followup}[1]{{\underline{#1}}}

\newcommand{\winnertable}[1]{{\textbf{#1}}}

\usepackage{amssymb}
\usepackage{pifont}
\newcommand{\cmark}{\ding{51}}%
\newcommand{\xmark}{\ding{55}}%

\definecolor{iccvblue}{rgb}{0.21,0.49,0.74}
\usepackage[pagebackref,breaklinks,colorlinks,allcolors=iccvblue]{hyperref}
\def\paperID{5295} 
\def\confName{ICCV}
\def\confYear{2025}

\usepackage{amsmath, amsfonts, times, epsfig, amssymb,  blindtext, float, subcaption, multicol, array, multirow, graphicx, etoolbox, tabularx, tabu, comment, soul}
\usepackage{grffile} 
 
\title{SignSplat: Rendering Sign Language via Gaussian Splatting}

\author{\parbox{16cm}{\centering
    {\large Maksym Ivashechkin, Oscar Mendez, Richard Bowden}\\
    {\normalsize
    CVSSP, University of Surrey, Guildford, United Kingdom}\\
    {\normalsize \texttt {\{m.ivashechkin, o.mendez, r.bowden\}@surrey.ac.uk}}}%
}

\begin{document}
\maketitle
\begin{abstract}
State-of-the-art approaches for conditional human body rendering via Gaussian splatting typically focus on simple body motions captured from many views.
This is often in the context of dancing or walking.
However, for more complex use cases, such as sign language, we care less about large body motion and more about subtle and complex motions of the hands and face. The problems of building high fidelity models are compounded by the complexity of capturing multi-view data of sign. 
The solution is to make better use of sequence data, ensuring that we can overcome the limited information from only a few views by exploiting temporal variability. Nevertheless, learning from sequence-level data requires extremely accurate and consistent model fitting to ensure that appearance is consistent across complex motions. 
We focus on how to achieve this, constraining mesh parameters to build an accurate Gaussian splatting framework from few views capable of modelling subtle human motion.
We leverage regularization techniques on the Gaussian parameters to mitigate overfitting and rendering artifacts.
Additionally, we propose a new adaptive control method to densify Gaussians and prune splat points on the mesh surface. 
To demonstrate the accuracy of our approach, we render novel sequences of sign language video, building on neural machine translation approaches to sign stitching.
On benchmark datasets, our approach achieves state-of-the-art performance; and on highly articulated and complex sign language motion, we significantly outperform competing approaches.
\end{abstract}

\section{Introduction}
\label{sec:intro}

Human body rendering has become a widely studied and dynamic area in computer vision, driven by the growing demand for realistic human representations in digital applications.
With advances in human-computer interaction, powerful graphic cards, and machine-learning techniques, the field is rapidly evolving, expanding its significance across a broad range of applications. 
Accurate and efficient rendering of human bodies is crucial for various tasks, including augmented/virtual reality, film production, gaming, and animation.

Our focus is accurate production of sign language where the intricate articulation of the human body, combined with nuanced facial expressions, demands a flexible human body model and a rendering method that is robust to complex deformations. Focusing on such an application highlights the deficiencies in current approaches. Furthermore, the common rejection of graphical avatars by the Deaf community means there is huge interest in approaches that employ generative AI to produce 2D photo realistic content~\cite{SaundersBen20200823}. However, modelling the human body in 3D, gives the animator far more flexibility.   

Within the literature, there are many 3D parameterizations for the human body. For example, implicit human body models such as imGHUM~\cite{Alldieck2021imGHUMIG} (also~\cite{peng2021neural, varol18_bodynet}, etc.) leverage the signed distance function (SDF) to capture intricate details and complex deformations which can result in a smooth representation.
Conversely, explicit (mesh-based) models, while offering efficient rendering, often suffer from artifacts due to their discrete nature and ability to represent fine-grained detail. Among the numerous mesh-based models \cite{STAR:2020,SMPL:2015,Anguelov_scale_2005}, etc.), SMPL-X~\cite{SMPL-X:2019} by Pavlakos~{\etal} is a popular choice, providing a detailed and expressive representation of the human body (e.g. \cite{Kato_2018_CVPR, ExPose:2020}).

Many approaches to differentiable rendering~\cite{Kanazawa_2018_cvpr} build upon these popular 3D body representations. 
However, they struggle with mesh artifacts and depend on the resolution of the mesh.
The implicit neural rendering methods such as Instant-NGP~\cite{mueller2022instant} or NeRF~\cite{mildenhall2020nerf} and derivates ({\eg} Plenoxels~\cite{yu2022plenoxels}, Mip-NeRF~\cite{Mip_NeRF}, etc.) have shown excellent capabilities in representing the scene as a continuous function, providing high-fidelity rendering quality, and allowing smooth interpolation between novel views.
The main disadvantages of neural renderers are the high computational cost needed to train the model and long inference times.
This was addressed in Gaussian splatting~\cite{kerbl3Dgaussians} of Kerbl~{\etal} who represent the scene as a set of discrete Gaussian points that enable real-time rendering with image quality often superior to that of neural renderers.
Nevertheless, even this approach can suffer from artifacts such as blur.

\subsection{Motivation}

The current benchmark datasets for human body rendering,~{\eg} \textsc{PeopleSnapshot}~\cite{alldieck2018video}, \textsc{NeuMan}~\cite{jiang2022neuman}, even~\textsc{X-Humans}~\cite{shen2023x}, etc. do not capture the full complexity of expressive human motion and gesture.
These datasets are good for overall human body evaluation.
However, they lack fine fidelity of motion such as expressive hand articulation, and this can result in~{\sota} methods overfitting to simple and broad human motions.



Moreover, the lack of articulation in benchmark evaluation data often leaves~{\sota} approaches underconstrained, leading to unrealistic artifacts and failures on challenging (and less represented) poses.
We tackle these challenges by developing a robust framework that combines single-view 3D reconstruction with Gaussian splatting rendering. To minimize rendering artifacts and enhance generalization while limiting data requirements, we regularize the Gaussian parameters and impose additional constraints on the underlying human representation.

Our goal is to leverage our framework to develop realistic sign language video where the robustness and accuracy of hand-to-hand interactions and motion are crucial to understanding. Even slight hand imprecision can alter the meaning in sign language. Existing approaches lack the explicit 3D priors and quality in rendering. As demonstrated in the experiments, our 3D human representation and efficient framework addresses these limitations.

\subsection{Contributions}
We present an SMPL-X-driven Gaussian splatting model for novel pose and novel view rendering that can be built from either single or multi-view inputs. We introduce the following key contributions:
\begin{enumerate}
    \item  A robust human representation for a Gaussian splatting model, capturing complex hand articulations.
    \item A pipeline for 3D uplift from 2D input with real-time image rendering via an efficient Gaussian splatting model.
    \item A novel 3D approach to sign-stitching for multi-gloss\footnote{A sign language gloss is a label that associates a word or words in English with a sign} rendering which enhances the smoothness and continuity of motion compared to existing 2D approaches.
\end{enumerate}

\noindent{}
We quantitatively compared our rendering model to the~{\sota} generic human body rendering methods, and qualitatively to the existing sign language production methods.
In both cases, we quantitatively and qualitatively outperform the current~{\sota}.

\begin{figure*}
    \centering
    \includegraphics[width=0.75\linewidth]{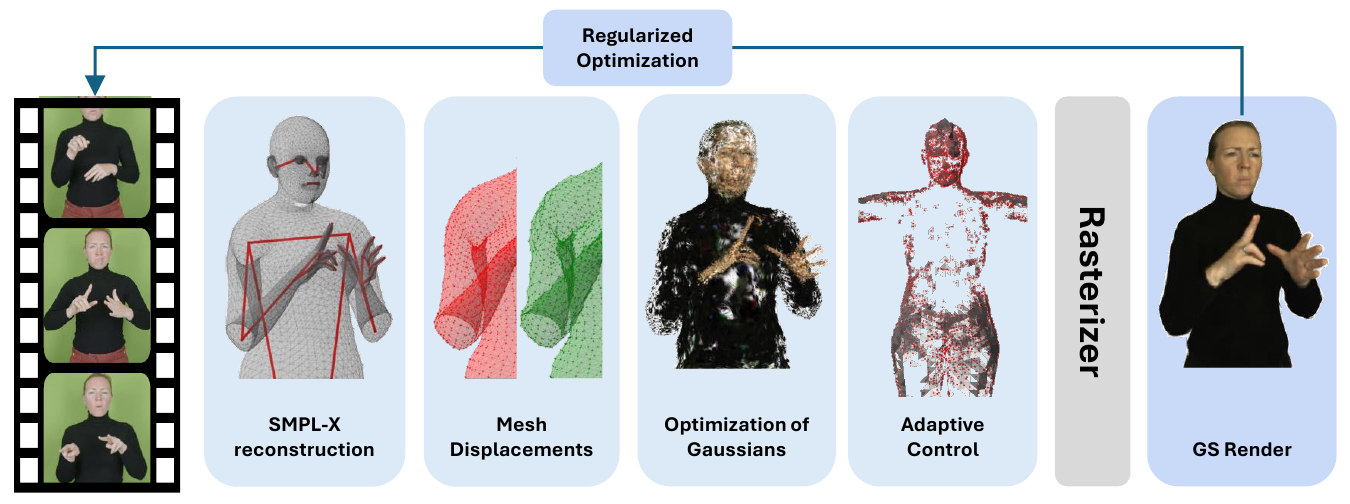}
    \vspace{-0.3cm}
    \caption{
    {\bf Overview:} The input is a set of images. First, we obtain SMPL-X mesh reconstruction to obtain an initial human mesh. Then, we iteratively optimize Gaussian parameters such as opacity, rotation, spherical harmonics coefficients, and scale, along with mesh displacements to refine the representation across all images. During this optimization process, we adaptively densify and prune points on the mesh surface, jointly regularizing both the mesh and Gaussians. 
    }
    \label{fig:pipeline-fig}
\end{figure*}
\section{Related Work}

The primary challenge in the human neural rendering field has been to integrate dynamic human articulation into rendering methods traditionally designed for static scenes.
A common approach to address this is to condition the renderer on mesh input, often a canonical pose, and learn deformation fields,~{\eg}~\cite{pumarola2020d,park2021hypernerf}, etc. that integrated deformations within a NeRF.

Early work focused on integrating human body models with NeRF.
For instance, A-NeRF~\cite{anerf} uses skeleton conditioning via a NeRF framework to render articulated humans.
Similarly, HumanNeRF~\cite{weng_humannerf_2022_cvpr} and 
NeuMan~\cite{jiang2022neuman} employ a canonical space for NeRF, based on the skeleton and SMPL~\cite{SMPL:2015} mesh parameters.
In a similar approach, Peng~{\etal}~\cite{Peng_animatable} introduces deformation fields within canonical space.
Neural Actor~\cite{liu2021neural} learns pose-dependent
geometric deformations and view-dependent appearances in
the canonical space, additionally leveraging texture maps.
H-NeRF~\cite{hnerf} by Xu~{\etal} leverages an implicit body representation via an SDF integrated with a neural radiance field.

Several papers propose approaches similar to NeRF neural rendering.
NeuralBody~\cite{peng2021neural} utilizes latent codes of a deformable mesh to encode geometry and appearance, while an MLP regresses the density and colors from the latent representation.
Later, Wang~{\etal} in~\cite{wang2023neuralrenderinghumansnovel} proposed a rendering method similar to NeuralBody that integrates observations across several frames and human appearance from input images.
Vid2Avatar~\cite{guo2023vid2avatar} decomposes scene reconstruction into both a human in canonical space and a separate background and provides SDF-based volume rendering.

However, Gaussian splatting~\cite{kerbl3Dgaussians} has risen in popularity due to it being significantly faster and more accurate than NeRF and its alternatives.
The current~{\sota} approaches are built on top of this.
SplattingAvatar~\cite{SplattingAvatar:CVPR2024} and similarly GaussianAvatars~\cite{qian2023gaussianavatars} define Gaussians as barycentric coordinates and displacements on a mesh.
HUGS~\cite{hugs} of Kocabas~{\etal} exploits a triplane feature map, where individual MLPs extract colors, geometry, and mesh linear blend-skinning (LBS) weights.
Similarly to the core functionality of HUGS, ExAvatar~\cite{moon2024exavatar} additionally uses pre-defined connectivity information to regularize the Gaussian splats.
GVA~\cite{liu24-GVA} reconstructs the human from a monocular view and constrains the surface of the avatar.

Many contributions have been proposed across the Gaussian splatting literature for human body rendering.
GaussianAvatar~\cite{hu2024gaussianavatar} uses a pose encoder from body points to feature space and a decoder to Gaussian parameters.  
To improve the quality of the SMPL mesh, which is commonly integrated into the Gaussian splatting framework, GST~\cite{prospero2024gstprecise3dhuman} authors train a transformer to predict small adjustments to the mesh.
GART~\cite{Lei_2024_CVPR} leverages the Gaussian Mixture model to extract appearance and shape in canonical space whereas Pang~{\etal} (in ASH~\cite{Pang_2024_CVPR}) propose to attach Gaussians to a deformable model and learn corresponding parameters in 2D texture space.
To address sparse views, GHG~\cite{kwon2024ghg} of Kwon~{\etal} also learns Gaussian parameters from the 2D UV space of the human template; and
additionally, the method constructs multi-scaffolds to effectively represent the offset details.
D3GA~\cite{Zielonka2023Drivable3G} moves from point to cage deformations.
GauHuman~\cite{gau_human} employs KL divergence to regulate the split, merge, and prune processes and Jiang~{\etal} (HiFi4G~\cite{Jiang_2024_CVPR}) integrate non-rigid tracking with a deformation graph, as well as creating efficient compressions of Gaussian points. 

In the sign-language domain, the literature mainly contains human generation from 2D keypoints.
Early approaches focused on avatar animation (\cite{Bangham_virtual,TESSA,EfthimiouE2012TdWE}, etc.), which were driven by glosses or expensive motion capture data.
More recent deep learning methods used for animation~\cite{SaundersBen20200823,SaundersBen2020ESNT} have exploited generative adversarial networks (GANs) to achieve more natural and realistic images.
However, these approaches heavily depend on 2D input skeleton detectors.
Walsh~{\etal}~\cite{SaundersBen20241125} leverage a 3D skeleton by stitching in 3D and then providing input to condition the GAN by projecting it to a rasterized 2D skeleton.
Despite achieving~{\sota} accuracy, the method heavily relies on the performance of the GAN for sign generation. 

The only application of Gaussian splatting to the domain of sign language we are aware of is the recently published EVA, by Hu~{\etal}~\cite{hu2024expressive} who build on the standard Gaussian splatting framework and apply it to Sign Language Sequences. As shown in the results, our framework outperform all approaches.


\section{Methodology}
The outline of our method is demonstrated in Fig.~\ref{fig:pipeline-fig}. A set of images, which can come from one or multiple cameras, is fed into our regularized optimization to create the Gaussian Splat model conditioned on an SMPL-X skeletal model. 

The proposed pipeline addresses the challenges posed by under-constrained and over-parameterized models, which are major contributors to failures in sign language production. For example,~{\sota} 3D rendering methods often overlook high articulation, while 2D-conditioned generative models lack 3D awareness. In the following sections, we present the key components that regularize and constrain the body model, adaptively incorporate finer details, and enhance image quality for continuous sign rendering.

\subsection{3D Reconstruction Pipeline}
Gaussian splatting, as well as all gradient descent optimization methods, are heavily dependent on initialization.
A suitable set of initial parameters leads to more accurate results and provides significantly faster convergence. Consequently, we take great care to obtain the initial reconstruction from the input 2D poses.

Vanilla Gaussian splatting methods typically rely on point cloud reconstruction from COLMAP~\cite{schoenberger2016sfm, schoenberger2016mvs} for initialization. However, we observed frequent failures for static or sparse camera setups, where COLMAP struggles to identify suitable image pairs. Instead, we use a parametric human model that provides a more robust point cloud representation.

For multiview data, we can perform an inverse kinematics reconstruction with SMPL-X parameters, 
by projecting the 3D model and minimizing the reprojection error with respect to 2D pose detections across all images. As the SMPL-X body model is a strong prior, we can also solve for camera extrincis in the same optimization. 

However, most sign-language datasets are single view. Therefore, to provide robust estimation we first segment the background using segment-anything~\cite{ravi2024sam2}, 2D keypoints are then detected via MMPose~\cite{mmpose2020} and we estimate the SMPL-X body with the OSX method~\cite{lin2023one}. We replace the predicted hands with a more accurate HaMer~\cite{pavlakos2024reconstructing} estimation.
These mesh parameters from the \emph{default} estimators are then further refined with our 2D re-projection fitting. We skin and project the 3D mesh back into the image and minimize the reprojection error to the 2D skeleton detections using approximated camera parameters.

Single-view reconstruction is fast, as the methods used run in real time and our 2D fitting converges quickly as we use PyTorch to run the gradient descent optimization. 
For 2D fitting, we deliberately use a low learning rate to prevent the initial reconstruction from deteriorating due to single-view ambiguities, as pre-trained models provide a strong human pose prior. However, fitting plays a crucial role in aligning the tentative 3D mesh with 2D detections.

Finally, we do mesh post-processing. The SMPL-X hand angular parameters consist of 45 values (3 rotation values for each of the 15 joints). However, 
the SMPL-X representation contains redundant degrees of freedom, which, in an under-constrained optimization, can lead to infeasible finger twists. If the optimization converges to a local minimum with unrealistic hand joint positions, Gaussian splatting alone cannot adequately resolve these errors.

To address this, we define feasible angular ranges and exclude certain rotation axes ({\ie} degrees of freedom) from the optimization process.
To efficiently enforce these constraints within the SMPL-X framework, which uses an axis-angle format by default, we leverage the Euler representation, as constraining angles is more intuitive and straightforward in this format.



\subsection{SMPL-X Human Body Representation}
SMPL-X is a unified differentiable human model consisting of body SMPL-H~\cite{SMPL:2015}, face FLAME~\cite{FLAME:SiggraphAsia2017}, and hand MANO~\cite{MANO:SIGGRAPHASIA:2017} parts. Formally, it takes  human shape coefficients ($\beta$), facial expressions ($\psi$), and  body angles ($\theta$). The SMPL-X model generates $N$ mesh points, where the underlying skeleton is extracted via linear blend skinning:
\begin{equation}
    M (\beta, \psi, \theta) \rightarrow \mathbb{R}^{3N}.
\end{equation}
The Gaussians are anchored to the mesh, with all Gaussian parameters set in a canonical space.
However, the positions of these points vary based on the SMPL-X parameters.

The original SMPL-X mesh resolution (number of points and edges) is sparse to provide high-fidelity rendering quality.
Consequently, we increase the mesh resolution by adding points at the center of large mesh faces and along long edges, based on a predefined threshold.
We observe that denser point clouds lead to significantly better results without deteriorating the efficacy and run-time of the Gaussian splatting rendering.
High-frequency details that can often be found on clothes and faces benefit from more points.


\subsection{Gaussian Splatting Parameters}

The main Gaussian splatting parameters anchored to the 3D points are colors (as spherical harmonics), anisotropic scales, rotations (quaternions), and opacities.
The Gaussian used in rendering is defined via a full 3D covariance matrix ($\Sigma = R S S^T R^T$) constructed from scale ($S$) and rotation ($R$), centered at position $\boldsymbol{\mu}$ and multiplied by opacity:
\begin{equation}
    G({\bf x}) = \text{exp}\big(-\frac{1}{2}{(\bf x}-\boldsymbol{\mu})^T\Sigma^{-1}{(\bf x} - \boldsymbol{\mu})\big).
\end{equation}

In original formulations, these parameters are optimized to fit a static 3D scene. However, for dynamic human motion, working with static sets of parameters is not desirable due to limb deformations, articulation, self-shadowing, etc.

In the literature, there is limited consensus on which Gaussian splatting parameters should be pose-dependent versus universal across a sequence. For instance, SplattingAvatar~\cite{SplattingAvatar:CVPR2024} and GaussianAvatars~\cite{qian2023gaussianavatars} use a single set of parameters, while other methods like ExAvatar~\cite{moon2024exavatar}, GaussianAvatar~\cite{hu2024gaussianavatar} or HUGS~\cite{hugs}, etc. predict some parameters, with others kept consistent throughout the sequence.

In our framework, for each Gaussian parameter, we design a shared light-weight 1D convolutional neural network $\Omega$.
It takes a point on a canonical mesh $\mathbf{x}_c$ and transformation from a new pose determined by the SMPL-X parameters $\mathcal{S}$ to a point on the canonical mesh. The $\Omega$ model predicts a feature embedding:
\begin{equation}
    \mathbf{f} = \Omega \big(\mathbf{x}_c, \phi(\mathbf{x}_c, \mathcal{S})\big).
\end{equation}

Afterward, four separate neural networks predict Gaussian attributes (scales, opacity, etc.) from this feature embedding $\mathbf{f}$.
The shared convolutional model learns human deformation changes while also leveraging the structure of neighboring points, and this means that the resulting output appears locally smoother and more consistent. The function $\phi$ returns a transform from observation pose to canonical. The transform is a translation between the canonical point and the current observation, which experimentally provided the best robustness to the model.

The beta coefficients of the mesh represent an approximate body shape.
However, they are not flexible enough to parameterize small details and body shape deformations.
Therefore, we leverage the displacements of the SMPL-X vertices along their normals.
We parameterize displacements of mesh vertices in 3D.
The displacements are also crucial to offset splat points when their Gaussian scale or rotation deforms. 
Let ${\bf v} \in \mathbb{R}^3$ be a mesh vertex, ${\bf n}_v \in \mathbb{R}^3$ be a vertex normal, then ${\bf d} \in \mathbb{R}^3$ is a point displacement in all axes, such that a new point position is:
\begin{equation}
\label{eq:displacements}
    {\bf v}^\prime = {\bf v} + {\bf d} \odot {\bf n}_v.
\end{equation}

The displacement parameters are predicted using another, similar, convolutional model, and the displacements are applied only to the original SMPL-X mesh vertices, excluding those that are densified later.

\subsection{Adaptive Density Control}
The densification process of Gaussian points plays an important role in Gaussian splatting optimization.
The authors of~\cite{kerbl3Dgaussians} implement cloning of the Gaussian points when they are too small to avoid under-reconstructed regions.
In contrast, when splat points are large, they are split into two smaller copies.
The decision to densify Gaussians depends on an average magnitude of view-space position gradients.

Unlike the SplattingAvatar~\cite{SplattingAvatar:CVPR2024} that updates vertices using a triangle-walking algorithm, we densify neighboring mesh faces to the selected point candidates.
If ${\bf x}$ is a point to be densified, then for $N$ adjacent faces, we create a new point to lie on a face by parameterizing it via 3 coefficients $k_1, k_2, k_3$ to create a convex combination ({\ie} $\sum_i^3 k_i = 1$) and a face normal offset $l$.
Therefore, for the verticies of a face $({\bf x}, {\bf y}, {\bf z})$ and face normal ${\bf n}_f$, the final position of a new point is:
\begin{equation}
    {\bf v} = k_1\cdot{\bf x} + k_2\cdot{\bf y}  + k_3\cdot{\bf z} + l\cdot{\bf n}_f.
\end{equation}

The proposed densification process aims to attach new points to the mesh, filling the gaps potentially caused by a low mesh resolution.
For each point, four new differentiable parameters are created and added into the optimization that can later adjust the initial location of new points.
We found that parameterizing points with an offset along the normal helps them to move inwards and upwards from the mesh surface, suitable for Gaussian scale offset.


Gaussian points are pruned if they become too large or if their opacity is close to zero. Since the points are attached to the mesh, they cannot be easily removed from the optimizer as in~\cite{kerbl3Dgaussians}; furthermore, some points are essential to maintain mesh integrity and consistency in neighboring areas.
As a result, we subsample points based on these criteria. Nevertheless, the original mesh points are retained during optimization to preserve the mesh structure.

\subsection{Constraints and Regularization}
A common discussion in the Gaussian splatting literature is the need for regularization to prevent rendering artifacts and to avoid convergence to a local minimum. 
ExAvatar~\cite{moon2024exavatar} penalizes the difference between the deformed and canonical mesh.
Lei~{\etal} in GART~\cite{Lei_2024_CVPR} minimize standard deviation on $k$-nearest neighbors.
SplattingAvatar~\cite{SplattingAvatar:CVPR2024} and GaussianAvatars~\cite{qian2023gaussianavatars} constrain the maximum scale and position of splats.
3DGS~\cite{qian20243dgs} apply as-isometric-as-possible regularization~\cite{Kilian_geometric}.
In the context of sign language, we differentiate between body, face, and hand parts, applying appropriate constraints and thresholds to each.

The first set of constraints we enforce are the hand parameters to remove redundant degrees of freedom.
Additionally, we set a maximum limit on the mesh vertex displacement ({\ie} magnitude $||{\bf d}||$ from the Eq.~\ref{eq:displacements}) along mesh normals considering hand, head, and body segments,~{\ie} allowing higher potential vertex displacement for the body and lower for the hands.

We define point neighborhoods based on mesh face connectivity.
Each neighboring region is also constrained by an overall radius, which considers both spatial proximity and Gaussian attribute similarity.
The idea is to minimize the variance of the Gaussian splatting parameters such as scales, rotations, colors, or opacity within the neighborhood to restrict body surface appearance.
This step ensures that densified splats are not significantly different within the mesh face.
Otherwise, high deviations in local areas lead to many artifacts for novel view and novel pose rendering. 
Importantly, we observed that spherical harmonics tend to overfit to training views, which leads to significant color inconsistency of the same Gaussian splat when rendered from different views. 
Enforcing the variance constraints potentially results in a trade-off, retaining high-frequency detail and overall consistency of Gaussian splats for novel views and poses.
This is mitigated by separating face points with more details from the body.
For each point neighborhood $\mathcal{N}$, the objective is variance minimization of the Gaussian set of parameters $\{{\bf p}\}$:
\begin{equation}
    \min \sum_{i\in\mathcal{N}} ({\bf p}_i - \frac{1}{|\mathcal{N}|}\sum_{i\in\mathcal{N}}{\bf p}_i)^2.
\end{equation}


To prevent Gaussian scales from expanding, we add a sigmoid layer to constrain the scales within pre-defined limits for the face, body, and hands.  
Additionally, to avoid overfitting to the training views, we keep the degrees of freedom for the spherical harmonics at zero for most iterations, gradually increasing them towards the end to fine-tune the model and include additional texture details.




\subsection{Sign Stitching}
We follow a gloss-based approach to Sign Stitching as proposed in~\cite{walsh-etal-2024-select} as this overcomes the data limitations of low-resource languages such as sign language. NMT-based translation approaches typically require over 15 million sequences of parallel data before they outperform statistical approaches~\cite{koehn-knowles-2017-six}. The largest single corpus of sign data consists of around 1.2M sentence pairs~\cite{albanie2021bbcoxfordbritishsignlanguage} which is around 1400 hours of data. Even the recently released multilingual dataset YouTube-SL-25~\cite{Tanzer2024YouTubeSL25AL}, which has over 3000 hours across 25 languages, remains an order of magnitude less than would be desirable. Following~\cite{walsh-etal-2024-select}, we use a large library of glosses captured in a multi-camera studio. We currently have a library of over 14,000 signs.
Converting an English sentence to a sign language sequence involves using part-of-speech (POS) tagging to identify candidate glosses in the library.
Reordering these to reflect sign grammar and then stitching the motions together using interpolation in the SMPL-X parameter space. 
 
To stitch individual glosses into a smooth continuous motion, we only use the last SMPL-X pose of the current gloss, and the first SMPL-X pose of the next gloss.
We convert the angular mesh (body and hands) parameters into quaternions and use spherical interpolation (SLERP).
However, instead of a linear step, we leverage a cubic ease-in ease-out animation formula that provides a gradual acceleration at the beginning of the motion and deceleration towards the end. 

The facial SMPL-X expression coefficients correspond to the zero-centered PCA values, so we use a simple linear interpolation. 
To avoid any jittering, we fix ({\ie} average over the sequence and keep constant) the global rotation and translation of the SMPL-X parameters.
Additionally, the beta body shape coefficients remain constant for the entire rendered sequence.

We adaptively determine the number of interpolated frames based on the range of motion. Specifically, we calculate the maximum angular difference between the final and first poses of the glosses sequence, then divide this value by a constant factor to obtain the appropriate frame count.

\section{Experiments}


Current~{\sota} methods perform well on datasets with minimal hand articulation and limited facial expressions. However, this is often due to over-parameterization and overfitting to much simpler motion patterns typical of some datasets (such as dancing or walking). Our body model incorporates adaptive control methods and additional regularization techniques to address these issues.

First we evaluate performance on multiple publicly available benchmark dataset and demonstrate~{\sota} performance. We then demonstrate the model's superior performance on sign-language data, where complex hand movements and expressions are essential. Finally, we showcase a novel mesh-based sign-stitching approach, followed by an analysis of the main components and an overall discussion of the results.

\subsection{Implementation Details}
We use the Adam optimizer~\cite{Kingma2014AdamAM} for training. For the objective function minimization, we exploit image reconstruction $\mathcal{L}_1$, D-SSIM, and LPIPS~\cite{zhang2018perceptual} losses with suitable weights.


In our framework, we jointly optimize all Gaussian parameters, mesh displacements, and SMPL-X mesh parameters. 
We observed that accumulating losses over multiple poses helps to stabilize the optimization process and significantly improves the overall accuracy.
This is because individual pose backpropagation pulls the Gaussian parameters towards a specific image to the detriment of other views.
However, when the losses are accumulated and then backpropagated, they average changes in the optimized tensors, thus regulating optimization.

\subsection{Quantitative Comparison}
First, we evaluate our method on the publicly available \textsc{NeuMan} dataset~\cite{jiang2022neuman}.
It includes short monocular images of a person captured ``in the wild''.
Following the common protocol ({\eg} from ExAvatar~\cite{moon2024exavatar}) we only evaluated our method on the \emph{bike}, \emph{jogging}, \emph{seattle}, and \emph{citron} sequences.
The comparison results are shown in Table \ref{table:neuman_no_bgr}.
As in previous works~\cite{chen2021animatable,qian20243dgs,hu2023gaussianavatar,moon2024exavatar,jiang2022instantavatar}, we fit SMPL-X parameters (excluding beta shape) and freeze the Gaussian parameters on testing frames using image loss.
We achieve the highest performance on all LPIPS, PSNR, and SSIM metrics.


\begin{table}[t]
\centering

\begin{tabular}{p{3.0cm}|>{\centering\arraybackslash}p{1.2cm}>{\centering\arraybackslash}p{1.2cm}>{\centering\arraybackslash}p{1.2cm}}
Methods & PSNR\textuparrow & SSIM\textuparrow & LPIPS\textdownarrow \\ \hline
HumanNeRF~\cite{weng2022humannerf} & 27.06 & 0.967 & 0.019 \\
InstantAvatar~\cite{jiang2023instantavatar} & 28.47 & 0.972 & 0.028 \\
NeuMan~\cite{jiang2022neuman} & 29.32 & 0.972 & 0.014 \\  
Vid2Avatar~\cite{guo2023vid2avatar} & 30.70 & 0.980 & 0.014 \\
GaussianAvatar~\cite{hu2023gaussianavatar} & 29.94 & 0.980 & 0.012 \\
3DGS-Avatar~\cite{qian20243dgs} & 28.99 & 0.974 & 0.016 \\
ExAvatar~\cite{moon2024exavatar} & 34.80 & 0.984 & 0.009 \\
Ours & \winnertable{35.47} &  \winnertable{0.986} & \winnertable{0.006}\\
\end{tabular}
\vspace{-0.2cm}
\caption{
Evaluation comparison on the test set of NeuMan~\cite{jiang2022neuman} dataset {\bf excluding} image backgrounds.}
\label{table:neuman_no_bgr}

\end{table}

We provide comparison results on the \textsc{X-Humans} dataset~\cite{shen2023xavatar}, which contains more expressive digital humans ({\eg} doing squats, moving hands) than the \textsc{NeuMan} dataset.  
To match the evaluation protocol, we use subsets \emph{00028}, \emph{00034}, and \emph{00087}, and the corresponding mesh parameters without post-training fine-tuning to test images.
The evaluation results are demonstrated in Table~\ref{table:xhumans}, where we outperform the~{\sota}.

\begin{table*}[t]
\centering
\setlength\tabcolsep{8pt}

\begin{tabular}{l|ccc|ccc|ccc}
\multirow{2}{*}{Methods} & \multicolumn{3}{c|}{\textit{\textbf{00028}}} & \multicolumn{3}{c|}{\textit{\textbf{00034}}} & \multicolumn{3}{c}{\textit{\textbf{00087}}} \\
 & PSNR\textuparrow & SSIM\textuparrow & LPIPS\textdownarrow & PSNR\textuparrow & SSIM\textuparrow & LPIPS\textdownarrow & PSNR\textuparrow & SSIM\textuparrow & LPIPS\textdownarrow\\ \hline
X-Avatar~\cite{shen2023x}$^*$ & 28.57 & 0.976 & 0.026 & 28.05 & 0.965 & 0.035 & 30.89  & 0.970 & 0.030 \\
ExAvatar~\cite{moon2024exavatar} & 30.58 & 0.981 & 0.018 & 28.75 & 0.966 & 0.029 & 32.01 & \textbf{0.972} & 0.025 \\
Ours &  {\bf 31.66} & {\bf 0.982} & {\bf 0.011} & {\bf 29.00} & {\bf 0.966} & {\bf 0.022} & {\bf 32.17} & {\bf 0.972} &  {\bf 0.022} \\

\end{tabular}
\vspace{-0.2cm}
\caption{
Evaluation on the test set of X-Humans~\cite{shen2023x} dataset. Label $^*$ denotes that competitor used additional depth maps for training.
}
\label{table:xhumans}

\end{table*}

\begin{figure}[t]
\centering
    \begin{subfigure}[t]{0.255\textwidth}
        \centering
        \includegraphics[width=0.7\linewidth]{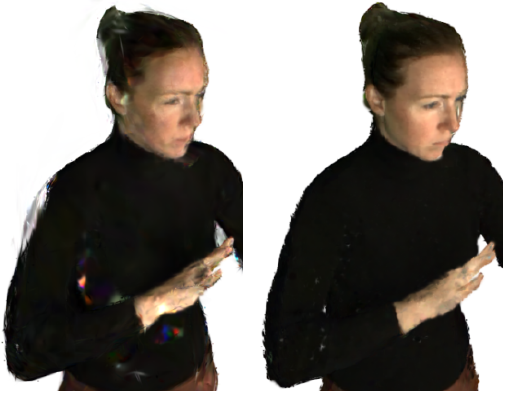}
        \caption{Regularization.}
        \label{fig:regularization_vis}
    \end{subfigure}%
    ~
    \begin{subfigure}[t]{0.215\textwidth}
        \centering
        \includegraphics[width=0.99\linewidth]{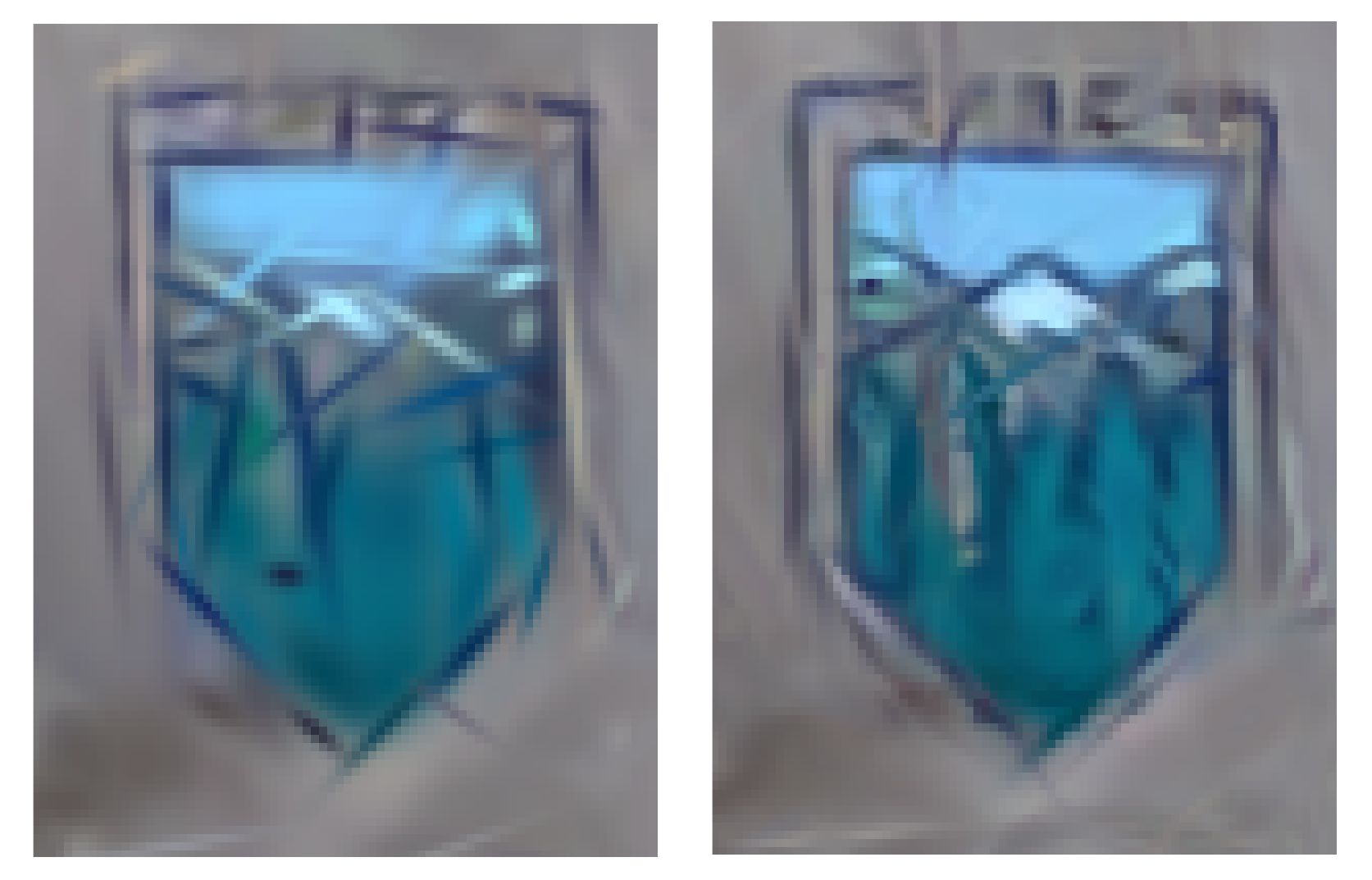}
        \caption{Adaptive density control.}
        \label{fig:densification_vis}
    \end{subfigure}
    \vspace{-0.2cm}
    \caption{Improvements on novel views (a) and novel poses (b).}
\end{figure}

We also compare the performance of our method on sign language data captured with only six views.
The recording contains a high level of articulation and facial expressions.
In total, we used 323 frames for training and 24 for testing that were sampled within a fixed interval.
We run competitors' code with the default setting provided, using instructions for video reconstruction. All approaches see the same data from all views. 
We estimated SMPL-X and camera parameters using our proposed single-view fitting framework.
The quantitative results on common metrics are shown in Table~\ref{table:rachel}, where we clearly outperform~{\sota} approaches by a large margin.

\begin{figure*}[t]
    \centering
    \includegraphics[width=0.76\linewidth]{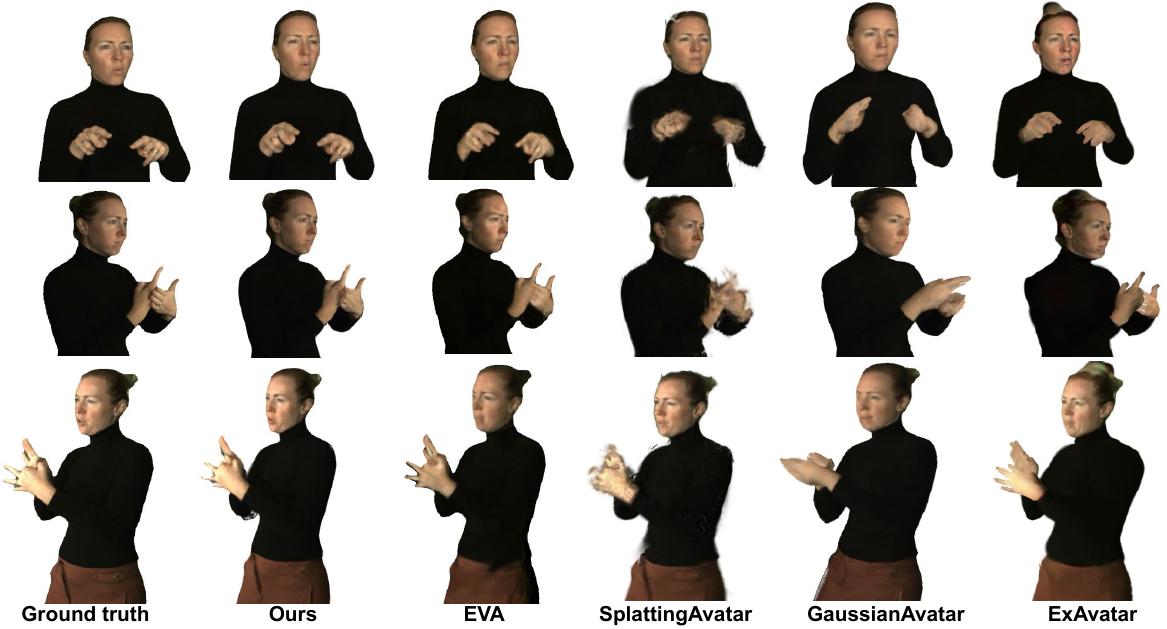}
    \vspace{-0.25cm}
    \caption{Qualitative comparison of rendering results with the~{\sota} on sign-language sequence.}
    \label{fig:rachel_qualitative}
\end{figure*}

\subsection{Qualitative Comparison}
\begin{figure*}
    \centering
    \includegraphics[width=0.75\linewidth]{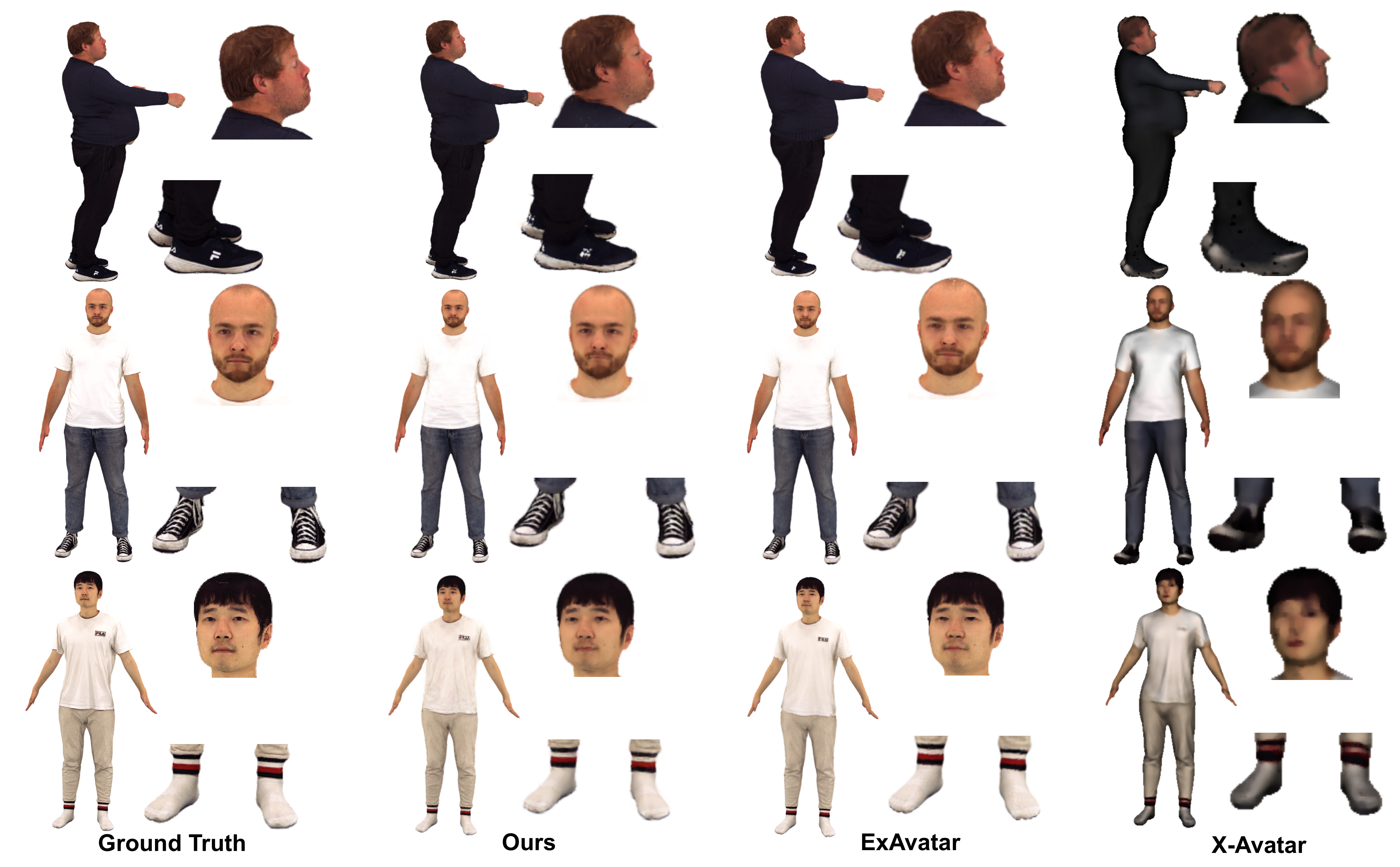}
    \vspace{-0.25cm}
    \caption{Qualitative comparison of rendering results with the~{\sota} on the~\textsc{X-Humans} dataset.}
    \label{fig:xhumans_vis}
\end{figure*}

We compare the visual performance of our method on the sign language data in Fig.\ref{fig:rachel_qualitative}. Our approach produces high-fidelity renderings that closely resemble ground-truth images. In contrast, the competitors struggle to render hands, and the hands look either oversmoothed or represented as blobs of Gaussian points. The closest approach to us in terms of performance is the recent EVA~\cite{hu2024expressive} method that also explicitly targets sign-language rendering. A visual comparison on the benchmark dataset \textsc{X-Humans} is shown in Fig.\ref{fig:xhumans_vis}, where our method similarly demonstrates high levels of detail in both clothing and facial features.


\subsection{Ablation Studies}

\begin{table}[t]
\centering
\begin{tabular}{p{3.0cm}|>{\centering\arraybackslash}p{1.2cm}>{\centering\arraybackslash}p{1.2cm}>{\centering\arraybackslash}p{1.2cm}}
Methods & PSNR\textuparrow & SSIM\textuparrow & LPIPS\textdownarrow \\ \hline
GaussianAvatar~\cite{hu2024gaussianavatar} & 23.873 & 0.968 & 0.032 \\
SplattingAvatar~\cite{SplattingAvatar:CVPR2024} & 32.313 & 0.985 & 0.016 \\
ExAvatar~\cite{moon2024exavatar} &  30.650 & 0.981 & 0.020 \\
EVA~\cite{hu2024expressive} &  33.193 & 0.975 & 0.013 \\
Ours & {\bf 33.769} & {\bf 0.989} & {\bf 0.007} \\

\end{tabular}
\vspace{-0.2cm}
\caption{Evaluation comparison on sign-language test set.}
\label{table:rachel}
\end{table}

\begin{table}[t]
    \centering
    \begin{tabular}{ccc|ccc}
        Reg. & Dens. & Acc. & PSNR\textuparrow & SSIM\textuparrow & LPIPS\textdownarrow  \\ \hline
        \xmark & \cmark & \cmark & 24.6035 & 0.9551 & \followup{0.0258} \\
        \cmark & \xmark & \cmark & \winner{24.8416}& \followup{0.9559} & 0.0287   \\
        \cmark & \cmark & \xmark &  24.0644 & 0.9528 & 0.0279 \\
        \cmark & \cmark & \cmark & \followup{24.6137} & \winner{0.9570}  & \winner{0.0249} \\
    \end{tabular}
    \vspace{-0.2cm}
    \caption{Impact of each part,~{\ie} ``Reg.'' -- regularization, ``Dens.'' -- adaptive control (densfication), ``Acc.'' -- accumulation of gradients, on the rendering accuracy.}
    \label{tab:ablations}
\end{table}


\begin{figure*}[t]
    \centering
    \includegraphics[width=0.9\linewidth]{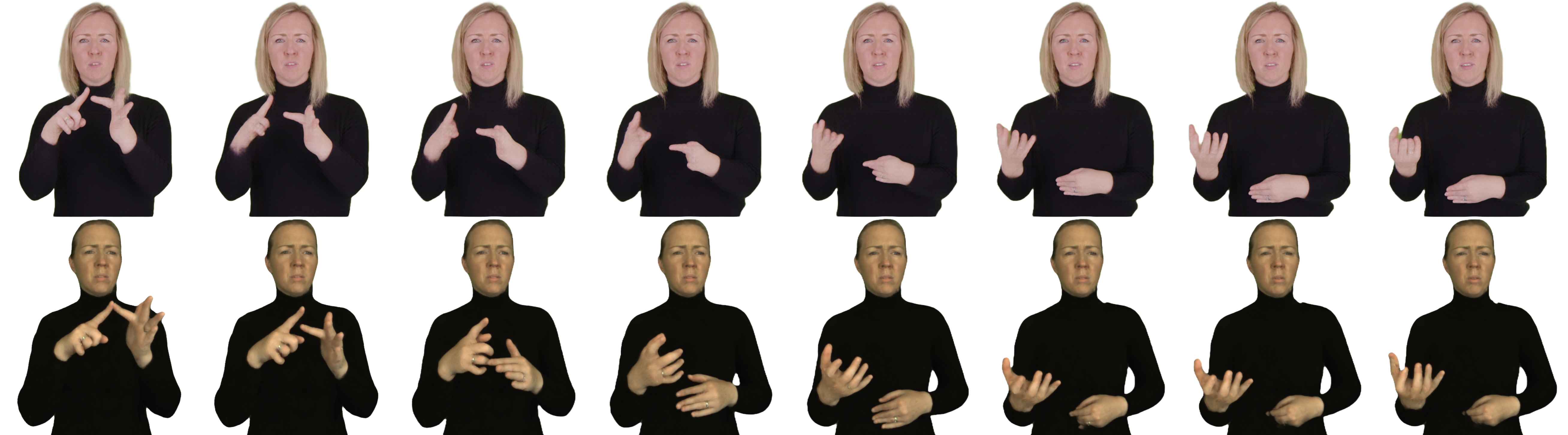}
    \vspace{-0.23cm}
    \caption{Comparison of sign gloss interpolation between a skeleton-based GAN approach and our rendering method.}
    \label{fig:sign_stitching}
\end{figure*}

We evaluate key features, including adaptive control and regularization, within the Gaussian splatting framework to empirically assess performance. The quantitative results, ran on the~\textsc{NeuMan} dataset, are presented in Table~\ref{tab:ablations}, where integrating all changes yields the best accuracy. Without regularization, novel views struggle with high scales and unexpected color variations, as shown in Fig.\ref{fig:regularization_vis}. Additionally, Fig.\ref{fig:densification_vis}
 demonstrates how adaptive control enhances the rendering of high-detail textures.

\subsection{Sign Stitching}

We compare our sign stitching approach against Saunders~{\etal}~\cite{SaundersBen2020ESNT} who interpolate 2D human poses, which are then fed into a GAN. However, many ambiguities arise in 2D, such as scale, shape, depth, etc. Interpolating mesh parameters in 3D helps to resolve most of these issues, ensuring a more consistent human shape. We present a comparison of this interpolation in Fig.~\ref{fig:sign_stitching}. Our approach produces solid and consistent hand rendering, whereas the fingers generated by the GAN tend to disappear or blend (see transition frames in the middle of the sequence and supplementary material).
\section{Conclusions}
We present a generic framework for novel pose and view rendering that targets highly expressive articulation and expressions, which is desirable for the applications such as sign language. 
The method is driven by a single-view SMPL-X mesh representing the human body, with canonical Gaussian parameters attached to the mesh to enable robust inference.
To ensure optimization convergence, we integrate suitable regularization on the Gaussian parameters and constrain the mesh.
We also propose a new adaptive densification control algorithm for cloning and pruning points on a mesh surface which enhances rendering quality.


We outperform~{\sota} on publicly available benchmark datasets. Nevertheless, most available datasets represent minimal hand articulation and focus on simple motions. To address this, we evaluate our method in the context of sign language where our method demonstrates superior rendering quality over current approaches. We also propose a novel mesh-based sign stitching approach for continuous sign language production.
Comparative evaluations with recent baselines demonstrate improved smoothness and continuity in articulation.

Visual artifacts are inevitable in any rendering method, and ours is no exception. To mitigate this issue, we regularize Gaussian splatting parameters such as scale and rotation to enforce consistency in the neighborhood structure. However, the effectiveness of this regularization often depends on factors such as the input data, human shape, and the number of views available. Moreover, excessive regularization can reduce accuracy and slow convergence. Therefore, adapting and optimizing parameter regularization remains crucial and could be further explored in future work.

{
    \small
    \bibliographystyle{ieeenat_fullname}
    \bibliography{references}
}


\end{document}